\documentclass[journal]{IEEEtran}
\usepackage{amsmath,amsfonts}
\usepackage{algorithmic}
\usepackage{algorithm}
\usepackage{array}
\usepackage{textcomp}
\usepackage{stfloats}
\usepackage{url}
\usepackage{verbatim}
\usepackage{graphicx}
\usepackage{cite}
\usepackage{balance}

\usepackage{amsmath,amsfonts,bm}

\def\eqref#1{equation~\ref{#1}}

\def\1{\bm{1}}

\DeclareMathAlphabet{\mathsfit}{\encodingdefault}{\sfdefault}{m}{sl}
\SetMathAlphabet{\mathsfit}{bold}{\encodingdefault}{\sfdefault}{bx}{n}

\def\gE{{\mathcal{E}}}

\def\gV{{\mathcal{V}}}

\usepackage{multirow}
\usepackage{microtype}
\usepackage{booktabs}
\usepackage{enumitem}
\usepackage[utf8]{inputenc}
\usepackage{listings}
\usepackage{caption}
\usepackage{dsfont}
\usepackage{graphicx}
\usepackage{subfig}
\usepackage{amsthm}
\usepackage{afterpage}
\usepackage{xspace}
\usepackage{adjustbox}

\usepackage{ragged2e} 
\usepackage{booktabs, tabularx}
\usepackage{makecell}
\usepackage{nicematrix,tikz}
\usepackage{wrapfig}

\usepackage{threeparttable}
\usepackage{hyperref}

\newcommand{\ours}{UGMAE\xspace}

\usepackage{arydshln}
\makeatletter
\def\adl@drawiv#1#2#3{%
        \hskip.5\tabcolsep
        \xleaders#3{#2.5\@tempdimb #1{1}#2.5\@tempdimb}%
                #2\z@ plus1fil minus1fil\relax
        \hskip.5\tabcolsep}
\newcommand{\cdashlineCustom}[1]{%
  \noalign{\vskip\aboverulesep
           \global\let\@dashdrawstore\adl@draw
           \global\let\adl@draw\adl@drawiv}
  \cdashline{#1}
  \noalign{\global\let\adl@draw\@dashdrawstore
           \vskip\belowrulesep}}
\makeatother

\begin{document}

\title{UGMAE: A Unified Framework for Graph Masked Autoencoders}

\author{Yijun Tian, Chuxu Zhang, Ziyi Kou, Zheyuan Liu, Xiangliang Zhang, Nitesh V. Chawla
\thanks{Yijun Tian is with the Department of Computer Science and Engineering, University of
Notre Dame, USA (e-mail: yijun.tian@nd.edu).}
\thanks{Chuxu Zhang is with the Department of Computer Science, Brandeis University, USA (e-mail: chuxuzhang@brandeis.edu).}
\thanks{Ziyi Kou is with the Department of Computer Science and Engineering, University of
Notre Dame, USA (e-mail: zkou@nd.edu).}
\thanks{Zheyuan Liu is with the Department of Computer Science and Engineering, University of
Notre Dame, USA (e-mail: zliu29@nd.edu).}
\thanks{Xiangliang Zhang is with the Department of Computer Science and Engineering, University of
Notre Dame, USA (e-mail: xzhang33@nd.edu).}
\thanks{Nitesh V. Chawla is with the Department of Computer Science and Engineering, University of
Notre Dame, USA (e-mail: nchawla@nd.edu).}
}

\maketitle

\begin{abstract}
Generative self-supervised learning on graphs, particularly graph masked autoencoders, has emerged as a popular learning paradigm and demonstrated its efficacy in handling non-Euclidean data. However, several remaining issues limit the capability of existing methods: 1) the disregard of uneven node significance in masking, 2) the underutilization of holistic graph information, 3) the ignorance of semantic knowledge in the representation space due to the exclusive use of reconstruction loss in the output space, and 4) the unstable reconstructions caused by the large volume of masked contents. In light of this, we propose \ours, a unified framework for graph masked autoencoders to address these issues from the perspectives of adaptivity, integrity, complementarity, and consistency. Specifically, we first develop an adaptive feature mask generator to account for the unique significance of nodes and sample informative masks (\textbf{adaptivity}). We then design a ranking-based structure reconstruction objective joint with feature reconstruction to capture holistic graph information and emphasize the topological proximity between neighbors (\textbf{integrity}). After that, we present a bootstrapping-based similarity module to encode the high-level semantic knowledge in the representation space, complementary to the low-level reconstruction in the output space (\textbf{complementarity}). Finally, we build a consistency assurance module to provide reconstruction objectives with extra stabilized consistency targets (\textbf{consistency}). Extensive experiments demonstrate that UGMAE outperforms both contrastive and generative state-of-the-art baselines on several tasks across multiple datasets.
\end{abstract}

\section{Introduction}

Although Graph Neural Networks (GNNs) have demonstrated exceptional effectiveness in handling graph data and have achieved great success in a wide range of graph mining tasks \cite{graphsage, gcn, gat, gnp, crowdgraph}, many of them adhere to the supervised or semisupervised learning settings \cite{semi_supervised_gnn_1, gnn_survey,semi_supervised_gnn_2, gnn_survey_2}, where labels are required to guide the learning procedure. To alleviate the strict requirements on labels, self-supervised learning on graphs is proposed and has emerged as one of the most exciting learning paradigms for GNNs \cite{ssl_graph_survey_1, ssl_graph_survey_2, hesgsl, g2gnn}. The key insight behind self-supervised learning on graphs is to learn node or graph representations based on supervision signals derived from the data itself without relying on human-annotated labels \cite{ssl_graph_survey_3, graphmixup}. The learned representations can then be utilized in various downstream applications and tasks such as recommendations \cite{ssl_recommendation_1, cml, mmssl} and molecular property prediction \cite{ssl_molecular_1, mol_survey}.

As for self-supervised learning on graphs, contrastive-based methods are one of the most prevalent and widely used approaches \cite{gcc, graphCL}. Since the insights behind contrastive learning are to maximize the agreement between contrastive data samples, the success of contrastive methods relies heavily on the derivation of these samples \cite{graphCL, contrastive_graph_1}. However, due to the fact that the majority of constructed data samples are based on heuristics and prior knowledge, model performance can vary between different tasks and data \cite{joao, contrastive_graph_2}. In addition, negative sampling, as a common strategy to obtain negative data samples for most of the contrastive objectives, often entails arduous designs and cumbersome constructions from graphs \cite{dgi, grace}. Therefore, deriving high-quality contrastive data samples becomes a crucial aspect of contrastive method designs.

Generative-based self-supervised learning methods naturally circumvent the aforementioned issues of building contrastive data samples by directly reconstructing the input graph data \cite{gae, gen_or_con, ssl_graph_survey_2}. Recently, masked autoencoders have exhibited remarkable learning capability in natural language processing \cite{bert} and computer vision \cite{mae} by eliminating a significant percentage of the input data and using the eliminated data to guide training. Inspired by this, some papers propose introducing masked autoencoders into the graph domain \cite{graphmae, gmae, mgae, maskgae, gigamae, hgmae}. For example, GraphMAE \cite{graphmae} and GraphMAE2 \cite{graphmae2} focuses on reconstructing node features with masking on graphs. GMAE \cite{gmae} leverages transformers as the encoder and decoder to capture the graph information. HGMAE \cite{hgmae} develops masked autoencoders on heterogeneous graphs. However, existing works have several limitations that hinder their capability in fully modeling the graph data and learning strong representations.

First, existing methods disregard the uneven node significance in masking. The random sampling of the masked nodes has shown to work, but since not all nodes contain the same amount of information, assuming a uniform probability distribution over all input nodes is sub-optimal. The second limitation is the underutilization of holistic graph information. Graphs are non-euclidean data with complex structural information and features. To learn effective node representations that fully encode the graph content and topology, merely reconstructing the edges or features is far from sufficient. In addition, current methods ignore the semantic knowledge in the representation space due to the exclusive use of reconstruction loss in the output space. Since the reconstruction loss focuses on the target in the output space while only the learned representations are utilized in the downstream tasks, a gap exists between the learning objective and what would be used ultimately. Moreover, masking a large portion of input data introduces implicit uncertainties into the model and causes unstable reconstructions, resulting in unsatisfactory model performance.

To address these challenges, we propose \ours, a novel unified framework for graph masked autoencoders, from the following four perspectives: adaptivity, integrity, complementarity, and consistency. Specifically, we first develop an adaptive feature mask generator to incorporate the node significance in masking. Compared to random masking, the designed adaptive masking generator samples more nodes that contain rich information and are hard to reconstruct. We then design a ranking-based structure reconstruction objective with the feature reconstruction loss to fully encode integral graph knowledge and emphasize the topological proximity between neighbor nodes. After that, a bootstrapping-based similarity module is proposed to complementarily integrate the high-level semantics from the latent space into the reconstruction in the output space. Finally, we build a consistency assurance module to promote stable learning by providing the original objectives with additional consistency labels. To fully evaluate our model, we conduct extensive experiments on three graph learning tasks. From the evaluation results, we conclude that \ours can effectively address the aforementioned challenges and achieve state-of-the-art performances. To summarize, the contributions of this paper are as follows:

\begin{itemize}[leftmargin=*]
    \item We point out that existing graph masked autoencoder methods suffer from four limitations that undermine their capability: the disregard of uneven node significance in masking, the underutilization of holistic graph information, the ignorance of semantic knowledge in the representation space, and the unstable reconstructions caused by the large volume of masked contents.

    \item To address these limitations, we propose \ours, a unified framework for graph masked autoencoders. \ours develops multiple novel components to tackle the issues from the perspectives of adaptivity, integrity, complementarity, and consistency. 

    \item Extensive experiments demonstrate the superiority of \ours compared to both contrastive and generative state-of-the-art baselines on three graph learning tasks.
\end{itemize}

\section{Related Work}

This work is closely related to graph neural networks, contrastive self-supervised learning on graphs and generative self-supervised learning on graphs.

\subsection{Graph Neural Networks}
Many GNNs are proposed to encode graph-structured data \cite{gat, gcn, graphsage, gin}. They take advantage of the message passing paradigm to learn representations for various graph learning tasks \cite{recipe2vec, gcf}. For example, GCN \cite{gcn} introduces a layer-wise propagation rule to encode node features. GAT \cite{gat} designs an attention mechanism for learning embeddings by aggregating information from neighbor nodes. Besides the node-level task, some GNNs are designed to achieve outstanding performances on graph-level tasks such as graph classification. For instance, GIN \cite{gin} proposes to model injective multiset functions for the neighbor aggregation and presents a simple structure that is as powerful as the Weisfeiler-Lehman graph isomorphism test. In addition, knowledge distillation \cite{hinton_kd} has been applied widely in graph-based research \cite{nosmog, kd_graph_survey}. Since knowledge distillation has shown great performance in enhancing GNN learning capacity \cite{gnnsd, freekd}, we introduce knowledge distillation in our model to ensure stable reconstructions.

\subsection{Contrastive Self-supervised Learning on Graphs}
Contrastive learning, which encourages alignment between different augmentations or distributions, has achieved great success in the graph domain and has become a widely recognized paradigm for graph representation learning \cite{infogcl, gcc, graphCL, joao, simgrace, datadec, reciperec, sgcl, engage}. For example, GCC \cite{gcc} aligns the local structures of two sampled subgraphs. GraphCL \cite{graphCL} focuses on aligning different graph augmentations. AD-GCL \cite{adgcl} introduces the alignment between the original graph and an adversarial augmented graph. However, the success of most contrastive methods relies on the choice of contrastive data samples \cite{graphCL, contrastive_graph_1}, where negative sampling is usually utilized. For instance, InfoGraph \cite{infograph} and DGI \cite{dgi} use corruptions to obtain negative pairs. GCC \cite{gcc} utilizes negative queues that are proposed in MoCo \cite{moco}. GraphCL \cite{graphCL}, GRACE \cite{grace}, and GCA \cite{contrastive_graph_1} leverage data samples within the same batch as negatives. Since the selection of different contrastive data samples can affect the performance greatly \cite{joao}, the determination and construction of suitable contrastive samples play an important role in contrastive method designs.

\subsection{Generative Self-supervised Learning on Graphs}
The objective of generative self-supervised learning on graphs is to reconstruct missing input data. In prior studies, the performance of generative methods is inferior to that of contrastive methods \cite{gae, gala, ep}. For example, the earliest works VGAE and GAE \cite{gae} leverage a graph convolutional network as the encoder and use a dot product operator as the decoder. Recently, GraphMAE \cite{graphmae} was proposed and has demonstrated great performance, which introduces masked autoencoder \cite{mae} in the graph domain by reconstructing node features, achieving outstanding performance. Simultaneously, many works are proposed to solve graph learning tasks following the pipeline of masked autoencoder. For instance, MGAE \cite{mgae} simply focuses on reconstructing the edges. GMAE \cite{gmae} uses transformers as the encoder and decoder instead of GNNs. MaskGAE \cite{maskgae} introduces path reconstruction and degree regression. HGMAE \cite{hgmae} develops graph masked autoencoders on heterogeneous graphs. However, existing methods disregard several challenges such as the ignorance of semantic knowledge in the representation space and the unstable reconstructions in the output space, which we address in this paper.

\begin{figure*}[t]
\begin{center}
\includegraphics[width=\textwidth]{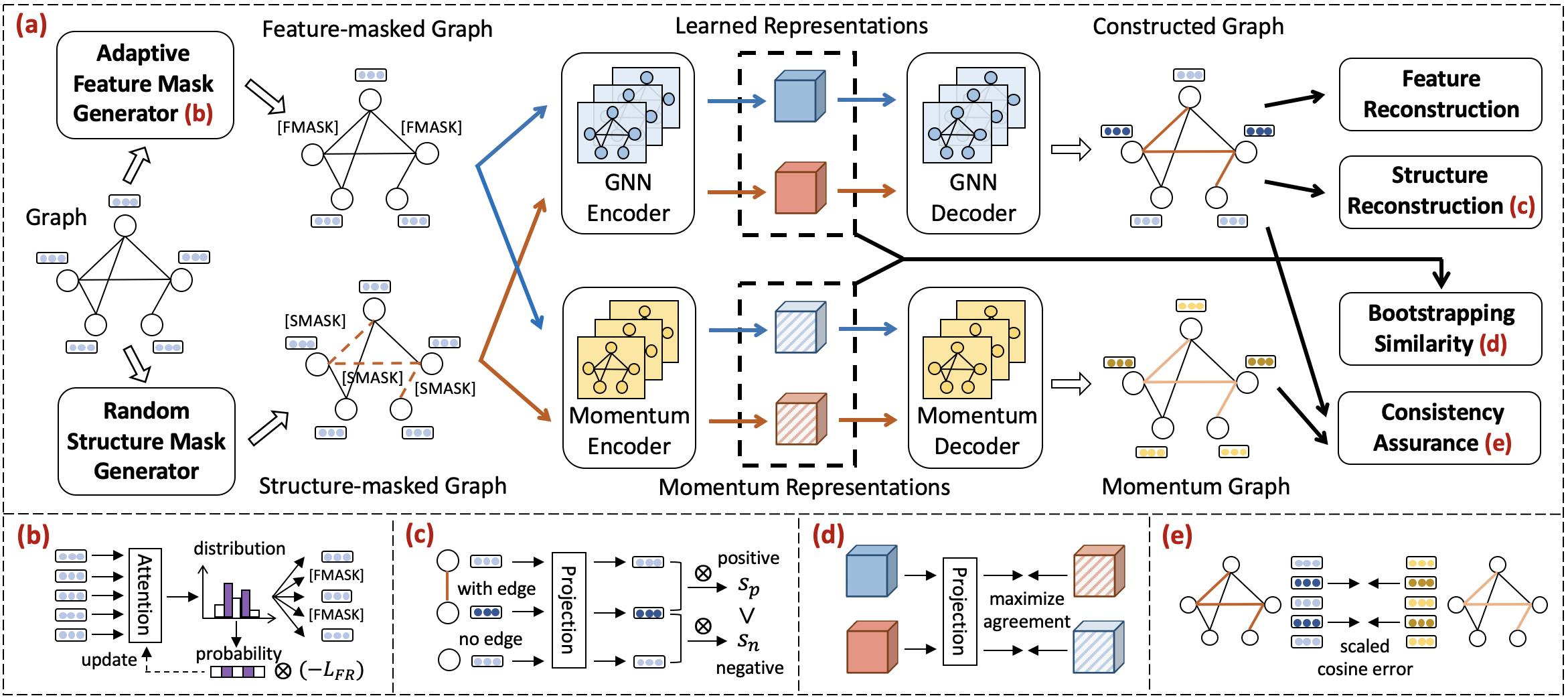}
\end{center}
\caption{
(a) The overall framework of \ours: we first design mask generators to obtain masked graphs and then send them into the encoder/decoder pipeline to learn representations. Then, the learned representations are taken for the reconstruction and consistency assurance objectives in the output space. In addition, we calculate bootstrapping similarity in the representation space to capture the high-level semantic knowledge.
(b) Adaptive feature mask generator: assigning high probability values for features with high reconstruction errors. 
(c) Ranking-based structure reconstruction: encouraging the connected nodes to be relatively similar.
(d) Bootstrapping-based similarity: maximizing the agreement between learned and momentum representations.
(e) Consistence assurance: minimizing the scaled cosine error between the constructed and momentum graph.
}
\label{fig:pipeline}
\end{figure*}

\section{Method}

In this section, we formally present \ours to address the limitations. As shown in Figure \ref{fig:pipeline} (a), \ours is a unified framework for graph masked autoencoders with multiple novel components, including an adaptive feature mask generator (Figure \ref{fig:pipeline} (b)), a ranking-based structure reconstruction objective (Figure \ref{fig:pipeline} (c)), a bootstrapping-based similarity module (Figure \ref{fig:pipeline} (d)), and a consistency assurance module (Figure \ref{fig:pipeline} (e)).

\subsection{Adaptive Feature Mask Generator}

In order to account for the unique significance of nodes in masking, we design the adaptive feature mask generator to obtain informative masks, as shown in Figure \ref{fig:pipeline} (b). By determining and assigning high probability values for node features with high reconstruction errors, the generator enables the model to focus on reconstructing nodes with hard features and rich information. In contrast to random sampling, we first utilize an auxiliary sampling network to estimate the categorical distribution for all nodes, and then obtain the masks based on the distribution. Given that sampling is a non-differentiable operation, we present a REINFORCE-based method to optimize the sampling network. In particular, we define the input graph as $G = (\gV, \gE, X)$, where $\gV$ represents the node set, $\gE$ indicates the edge set, and $X$ denotes the node features. To calculate and enable the model to consider the node significance, we first use a multi-head attention network (MHA) followed by a simple feed-forward neural network (FFN) and a Softmax activation to calculate the probability scores $P = \text{Softmax} \left ( \text{FFN} [ \text{MHA} (X)] \right )$.

We then define a categorical distribution over $P$ and draw the set of masked nodes $\widetilde{\gV}$ without replacement, where $\widetilde{\gV} \in \gV$. The number of nodes in $\widetilde{\gV}$ is determined by the feature mask rate $p_f$, where $|\widetilde{\gV}| = |\gV| \times p_f$. After that, We design a learnable mask token $[\text{FMASK}]$ to mask the node features. For each node $v \in \widetilde{\gV}$, we have its mask feature $X_{[\text{FMASK}]} \in \widetilde{X}$, where $\widetilde{X}$ denotes the masked feature matrix. Correspondingly, for each $\widetilde{X}^{v} \in \widetilde{X}$, the process is formulated as follows:
\begin{equation}
    \widetilde{X}^{v}=
    \begin{cases}
    X_{[\text{FMASK}]} \quad & \text{if $v\in\widetilde{\gV}$} \\
    X^v \quad & \text{if $v\notin\widetilde{\gV}$}. 
    \end{cases}
\end{equation}

\noindent
\textbf{Feature Reconstruction.}
After we acquire the masked feature matrix $\widetilde{X}$, we denote the feature-masked graph as $\widetilde{G}_f$ where $\widetilde{G}_f = (\gV, \gE, \widetilde{X})$, which is later send into the encoder $f_E$ to obtain the learned node representations $H_1$. To encourage the encoder to learn informative representations without relying on the decoder's capability for reconstruction, we apply another mask token $[DM]$ on $H_1$ to obtain $\widetilde{H}_1$ before sending it into the decoder, where $[DM]$ holds the same masked node indices as $[FMASK]$. Next, we send $\gE$ and $\widetilde{H}_1$ into the decoder $f_D$ to obtain the reconstructed node features $Z_1$ under feature masking. The process is as follows:
\begin{equation}
    \begin{split}
        H_1 = f_E(\gE, \widetilde{X}),
        \quad
        \widetilde{H}_1=
        \begin{cases}
        h_{[DM]} \quad & \text{if $v\in\widetilde{\gV}$} \\
        h^v \quad & \text{if $v\notin\widetilde{\gV}$}, 
        \end{cases} 
        \quad
        Z_1 = f_D(\gE, \widetilde{H}_1).
    \end{split}
\end{equation}
Subsequently, for each node $v \in \widetilde{\gV}$, we compute the loss between the reconstruction $Z_1^v \in Z_1$ and the ground-truth node features $X^v \in X$ with a scaling factor $\alpha$. In particular, we define the feature reconstruction loss $\mathcal{L}_{\textrm{FR}}$ as follows:
\begin{equation}
    \mathcal{L}_{\textrm{FR}} = \frac{1}{|\widetilde{\gV}|} \sum_{v \in \widetilde{\gV}} (1 - \frac{X^v \cdot Z_1^v}{\| X^v \| \times \| Z_1^v\|})^{\alpha}.
\end{equation}

\noindent
\textbf{Optimizing Adaptive Feature Mask Generator.}
Since the sampling process is non-differentiable, to optimize the adaptive feature mask generator, we present a REINFORCE-based method. Specifically, we introduce a sampling loss $\mathcal{L}_{\textrm{sample}}$, in which we consider the sampling process as \textit{action}, both the encoder and the decoder as \textit{environment}, and the feature reconstruction loss $\mathcal{L}_{\textrm{FR}}$ as \textit{return}. To encourage the generator to focus on nodes that are hard to reconstruct, we optimize the generator by maximizing the expected feature reconstruction error $\mathbb{E}[\mathcal{L}_{FR}]$: $\mathcal{L}_{\textrm{sample}} = - \mathbb{E}[\mathcal{L}_{\textrm{FR}}] =  - \sum_{v \in \widetilde{\gV}} log(P^v) \cdot \mathcal{L}_{\textrm{FR}}^v$, where $P^v$ is the probability score for node $v$ and  $\mathcal{L}_{\textrm{FR}}^v$ is the feature reconstruction error for node $v$. To prevent precision errors caused by small values, we take the logarithm for $P^v$. In addition, we stop the gradient updates from $\mathcal{L}_{\textrm{sample}}$ to propagate through the encoder and decoder with the aim of avoiding duplicate computation.

\subsection{Ranking-based Structure Reconstruction}

We also design a structure reconstruction objective to capture holistic graph knowledge and emphasize the topological proximity between neighbors, as shown in Figure \ref{fig:pipeline} (c). Since reconstructing the edges via a strict binary classification loss might force the model to focus on the explicit structures while ignoring the implicit relevance between nodes, we design a ranking-based objective to incorporate the relative node similarities and distances. The objective involves comparing the preference scores between nodes connected by an edge against the scores between nodes without any connections. To demonstrate, we start by introducing a random structure mask generator to obtain structure mask $[SMASK]$. Specifically, we randomly sample a subset of masked edges $\gE_{mask} \in \gE$ following Bernoulli distribution, i.e., $\gE_{mask} \sim \text{Bernoulli}(p_s)$, where $p_s$ denotes the structure mask rate. For edges in $\gE_{mask}$, we mask them with $[SMASK]$, deriving the structure-masked graph $\widetilde{G}_s = (\gV, \widetilde{\gE}, X)$, where $\widetilde{\gE}$ represents the remained visible edges after masking with $\widetilde{\gE} = \gE-\gE_{mask}$. Then, we feed $\widetilde{\gE}$ and node features $X$ into the encoder $f_E$ to obtain the learned node representations $H_2$ under structure masking. After that, we send $\widetilde{\gE}$ and $H_2$ into the decoder $f_D$ to generate the reconstructed node features $Z_2$ under structure masking. The process is formulated as follows:
\begin{equation}
    \begin{split}
        H_2 = f_E(\widetilde{\gE}, X),
        \quad
        Z_2 = f_D(\widetilde{\gE}, H_2).
    \end{split}
\end{equation}
After we obtain $Z_2$, we can leverage it to calculate the preference scores between nodes, and further use the scores to define the structure reconstruction objective. To illustrate, given two connected nodes $v_i$ and $v_j$, we encourage the inner product similarity between $v_i$ and $v_j$ to be higher than the inner product similarity between $v_i$ and a random negative node $v_{j'}$. Accordingly, we formulate the structure reconstruction loss $\mathcal{L}_{\textrm{SR}}$ as follows:
\begin{equation}
\mathcal{L}_{\textrm{SR}} = \sum_{(v_i, v_j) \in \widetilde{\gE}} \text{max}\{0, 1-\text{sim}(Z_2^{i}, Z_2^{j})+\text{sim}(Z_2^{i}, Z_2^{j'})\},
\end{equation}
where $Z_2^{i}, Z_2^{j}, Z_2^{j'}$ are learned representations under structure masking for $v_i, v_j, v_{j'}$, respectively.

\subsection{Bootstrapping-based Similarity}

We propose a bootstrapping-based similarity module (Figure \ref{fig:pipeline} (d)) to capture high-level semantic knowledge, which enables the model to iteratively bootstrap the prior encoder outputs as learning targets and facilitate enhanced representations. In particular, we refine the learned representations via a bootstrapping procedure by predicting the output of a momentum encoder, i.e., predicting the momentum representations that are generated by a slowly moving exponential average of the original encoder. We start by introducing a momentum encoder $f_{E^*}$ with the same model structure as the original encoder $f_{E}$. We set the parameters for $f_{E^*}$ as an updated exponential moving average of $f_{E}$. We then feed the feature-masked graph $\widetilde{G}_f = (\gV, \gE, \widetilde{X})$ and the structure-masked graph $\widetilde{G}_s = (\gV, \widetilde{\gE}, X)$ into $f_{E^*}$ to obtain the momentum representations $H_1^*$ and $H_2^*$, repestively. The process is formulated as follows:
\begin{equation}
    \label{eq:H1_star}
    \begin{split}
        H_1^* = f_{E^*}(\gE, \widetilde{X}),
        \quad
        H_2^* = f_{E^*}(\widetilde{\gE}, X).
    \end{split}
\end{equation}
As momentum representations, $H_1^*$ and $H_2^*$ provide dynamically deeper semantics by considering prior knowledge via bootstrapping. To enable better learning, we transform the learned representations $H_1$ and $H_2$ via a shared simple projection network $\text{proj}$ (e.g., MLP):
\begin{equation}
    \begin{split}
        H_1' = \text{proj}(H_1),
        \quad
        H_2' = \text{proj}(H_2),
    \end{split}
\end{equation}
where $H_1'$, $H_2'$ are the transformed representations of $H_1$, $H_2$, respectively. Next, we encourage the cross-masking similarity by making $H_1'$ on $\widetilde{G}_f$ closer to $H_2^*$ on $\widetilde{G}_s$, and $H_2'$ on $\widetilde{G}_s$ closer to $H_1^*$ on $\widetilde{G}_f$. Given that the momentum representations are generated via the bootstrapping procedure, we name the similarity as the bootstrapping-based similarity. We promote this similarity for each node $v \in \gV$ and formally define the loss $\mathcal{L}_{\textrm{BS}}$ as follows:
\begin{equation}
    \mathcal{L}_{\textrm{BS}} = -\frac{1}{|\gV|} \sum_{v \in \gV} 
    \left (
    \frac{(H_1')^v \cdot (H_2^*)^v}{\| (H_1')^v \| \times \| (H_2^*)^v \|}
    +
    \frac{(H_1^*)^v \cdot (H_2')^v}{\| (H_1^*)^v \| \times \| (H_2')^v \|}
    \right ).
    \label{eq:loss_bs}
\end{equation}

\subsection{Consistency Assurance}

With the aim of alleviating the inconsistent reconstruction caused by the large portion of different masked contents, we design a consistency assurance module to provide the model with extra stabilized consistency targets, as shown in Figure \ref{fig:pipeline} (e). To demonstrate, we introduce a self-distillation learning paradigm by distilling the knowledge from a momentum decoder teacher to the original decoder student, with the aim of improving the student by learning from the teacher. In particular, we force the original decoder student to learn from the momentum decoder teacher by matching the reconstructions. We find that the momentum decoder teacher functions in the form of model ensembling similar to Polyak-Ruppert averaging with exponential decay, and utilizing Polyak-Ruppert averaging for model ensembling is a standard way for enhancing model capability \cite{polyak, polyak_for_ensembl}, which further enables the original student decoder to learn and develop high-quality reconstructions. In specific, we denote the momentum decoder teacher as $f_{D^*}$, and equip it with the same model structure as the original decoder student $f_{D}$. The parameters of $f_{D^*}$ are updated by the exponential moving average of $f_{D}$. Then, we send the edge set $\gE$ and the momentum representations $H_1^*$ (from Eq. \ref{eq:H1_star}) into $f_{D^*}$ to obtain the momentum reconstructions $Z_1^*$. The process is shown as follows:
\begin{equation}
    Z_1^* = f_{D^*}(\gE, H_1^*).
\end{equation}
Next, we compare the original reconstructions $Z_1$ and the momentum reconstructions $Z_1^*$, and further encourage them to be similar via a scaled cosine error. Since we stop the gradient from the momentum decoder, $Z_1$ is forced to learn from $Z_1^*$. We employ the consistency assurance for every node $v \in \widetilde{\gV}$ and define the consistency assurance loss $\mathcal{L}_{\textrm{CA}}$ as follows:
\begin{equation}
    \mathcal{L}_{\textrm{CA}} = \frac{1}{|\widetilde{\gV}|} \sum_{v \in \widetilde{\gV}} (1 - \frac{Z_1^v \cdot (Z_1^*)^v}{\| Z_1^v\| \times \| (Z_1^*)^v \|})^{\beta},
\end{equation}
where $\beta$ is the scaling factor. Theoretically, if $(Z_1^*)^v$ and $Z_1^v$ are very similar, then $\mathcal{L}_{\textrm{CA}}$ will be small, encouraging the model to learn more from the reconstruction objectives. However, if $(Z_1^*)^v$ and $Z_1^v$ differ a lot, then $\mathcal{L}_{\textrm{CA}}$ will be large, facilitating the model to move in a direction that favors consistency provided by prior knowledge. In addition, the utilization of the scaling factor $\beta$ enables the decoder to down-weight the contributions of similar reconstruction pairs when $\beta > 1$ while focusing more on those pairs that are distinct. Finally, we train the model by combining the above objectives.

\begin{table*}[htbp]
    \centering
    \caption{Node classification performance comparison. We report accuracy (\%) for all datasets except Micro-F1 (\%) score for PPI dataset. The best and second-best results are highlighted in bold and underlined, respectively. 
    }
    \renewcommand\tabcolsep{11.5pt}
    \renewcommand\arraystretch{1.05}
    \resizebox{\linewidth}{!}{
    \begin{NiceTabular}{c|c|ccccc}
        \toprule[1.1pt]
        Types    & Methods &   Cora      & CiteSeer      & PubMed                & Ogbn-arxiv        & PPI     \\
         \midrule
        \multirow{2}{*}{Supervised} 
        & GCN \cite{gcn}    &  81.5          & 70.3          & 79.0                   & 71.74$\pm$0.29    & 75.7$\pm$0.1      \\
        & GAT \cite{gat}   &  83.0$\pm$0.7  & 72.5$\pm$0.7  & 79.0$\pm$0.3           & 72.10$\pm$0.13     & 97.30$\pm$0.20      \\
        \midrule
        \multirow{11.5}{*}{Self-supervised} 
        & GAE \cite{gae}    &  71.5$\pm$0.4  & 65.8$\pm$0.4  & 72.1$\pm$0.5           & -               & -        \\
        & GPT-GNN \cite{gptgnn} &  80.1$\pm$1.0  & 68.4$\pm$1.6  & 76.3$\pm$0.8 & - & - \\
        & GATE \cite{gate}   &  83.2$\pm$0.6  & 71.8$\pm$0.8  & 80.9$\pm$0.3           & -                 & -     \\ 
        & DGI \cite{dgi}    &  82.3$\pm$0.6  & 71.8$\pm$0.7  & 76.8$\pm$0.6           & 69.68$\pm$0.13 & 63.80$\pm$0.20  \\
        & MVGRL \cite{mvgrl}  & 83.5$\pm$0.4   & 73.3$\pm$0.5  & 80.1$\pm$0.7           & -               & -  \\
        & GRACE \cite{grace}  & 81.9$\pm$0.4   & 71.2$\pm$0.5  & 80.6$\pm$0.4           & - & 69.71$\pm$0.17 \\  
        & BGRL \cite{bgrl}    & 82.7$\pm$0.6   & 71.1$\pm$0.8  & 79.6$\pm$0.5           & 70.30$\pm$0.14   & 73.63$\pm$0.16     \\
        & InfoGCL \cite{infogcl}  & 83.5$\pm$0.3   & \underline{73.5$\pm$0.4}  & 79.1$\pm$0.2  & - & - \\
        & CCA-SSG \cite{ccassg} & 84.0$\pm$0.4   & 73.1$\pm$0.3  & 81.0$\pm$0.4  & 70.16$\pm$0.22  & 73.34$\pm$0.17   \\
        & GraphMAE \cite{graphmae} & 84.2±0.4  & 73.4±0.4  & 81.1±0.4  & 70.37$\pm$0.11 & 74.50$\pm$0.29    \\
        & GraphMAE2 \cite{graphmae2} & \underline{84.4±0.5}  & 73.4±0.3  & \underline{81.4±0.5}  & \underline{70.43$\pm$0.24} & \underline{74.51$\pm$0.36}    \\
        \cmidrule{2-7}
        & \ours  & \bf 85.1$\pm$0.4 & \bf 73.9$\pm$0.3 & \bf 82.2$\pm$0.1 & \bf 70.60$\pm$0.06 & \bf 74.83$\pm$0.33 \\
        \bottomrule[1.1pt]
    \end{NiceTabular}
    }
    \label{tab:node_classification}
\end{table*}

\begin{table*}[htbp]
    \centering
    \caption{Graph classification performance comparison. We report accuracy (\%) for all datasets.
    }
    \renewcommand\arraystretch{1.1}
    \resizebox{\linewidth}{!}{
    \begin{NiceTabular}{c|c|ccccccc}
        \toprule[1.1pt]
        Types      & Methods  & IMDB-M     & IMDB-B     & PROTEINS   & COLLAB     & MUTAG      & REDDIT-B   & NCI1     \\
        \midrule
        \multirow{2}{*}{Supervised}
        & GIN \cite{gin}        & 52.3$\pm$2.8   & 75.1$\pm$5.1   & 76.2$\pm$2.8   & 80.2$\pm$1.9   & 89.4$\pm$5.6   & 92.4$\pm$2.5   & 82.7$\pm$1.7 \\
        & DiffPool \cite{diffpool}   &  -           & 72.6$\pm$3.9 &  75.1$\pm$3.5   & 78.9$\pm$2.3 & 85.0$\pm$10.3 & 92.1$\pm$2.6 & - \\
        \midrule
        \multirow{2}{*}{Graph Kernels}
        & WL  \cite{wl}        & 46.95$\pm$0.46 & 72.30$\pm$3.44 & 72.92$\pm$0.56 & - & 80.72$\pm$3.00 & 68.82$\pm$0.41 & 80.31$\pm$0.46 \\
        & DGK  \cite{dgk_and_graph_classification_datasets}       & 44.55$\pm$0.52 & 66.96$\pm$0.56 & 73.30$\pm$0.82 & - & 87.44$\pm$2.72 & 78.04$\pm$0.39 & 80.31$\pm$0.46 \\
        \midrule
        \multirow{9.5}{*}{Self-supervised}
        & graph2vec \cite{graph2vec}  & 50.44$\pm$0.87 & 71.10$\pm$0.54 & 73.30$\pm$2.05 & -              & 83.15$\pm$9.25 & 75.78$\pm$1.03 & 73.22$\pm$1.81 \\
        & Infograph \cite{infograph}  & 49.69$\pm$0.53 & 73.03$\pm$0.87 & 74.44$\pm$0.31 & 70.65$\pm$1.13 & 89.01$\pm$1.13 & 82.50$\pm$1.42 & 76.20$\pm$1.06 \\
        & GraphCL \cite{graphCL}    & 48.58$\pm$0.67 & 71.14$\pm$0.44 & 74.39$\pm$0.45 & 71.36$\pm$1.15 & 86.80$\pm$1.34 & 89.53$\pm$0.84 & 77.87$\pm$0.41 \\
        & JOAO \cite{joao}       & 49.20$\pm$0.77 & 70.21$\pm$3.08     & 74.55$\pm$0.41 & 69.50$\pm$0.36 & 87.35$\pm$1.02 & 85.29$\pm$1.35 & 78.07$\pm$0.47 \\
        & GCC  \cite{gcc}       & 49.4           & 72.0           & -    & 78.9    &  - & \bf 89.8 & - \\
        & MVGRL \cite{mvgrl}      & 51.20$\pm$0.50 & 74.20$\pm$0.70     & -              & -              & \underline{89.70$\pm$1.10}   & 84.50$\pm$0.60   & -               \\
        & InfoGCL \cite{infogcl}    & 51.40$\pm$0.80   & 75.10$\pm$0.90   &  -             & 80.00$\pm$1.30   & \bf 91.20$\pm$1.30   & -              &  80.20$\pm$0.60   \\
        & GraphMAE \cite{graphmae}    & 51.63$\pm$0.52 & \underline{75.52$\pm$0.66} & \underline{75.30$\pm$0.39} & \underline{80.32$\pm$0.46} & 88.19$\pm$1.26 & 88.01$\pm$0.19 & \underline{80.40$\pm$0.30}  \\
        & GraphMAE2 \cite{graphmae2}    & \underline{51.80$\pm$0.60} & 73.88$\pm$0.53 & 74.86$\pm$0.34 & 77.59$\pm$0.22 & 86.63$\pm$1.33 & 76.84$\pm$0.21 &  78.56$\pm$0.26 \\
        \cmidrule{2-9}
        & \ours  & \bf 52.64$\pm$0.30 & \bf 76.06$\pm$0.59 & \bf 76.78$\pm$0.22 & \bf 81.66$\pm$0.12 & 88.26$\pm$1.19 & \underline{89.54$\pm$0.21} & \bf 80.46$\pm$0.17 \\
        \bottomrule[1.1pt]
    \end{NiceTabular}
    }
    \label{tab:graph_classification}
\end{table*}

\section{Experiments}

In this section, we conduct extensive experiments to validate the effectiveness and applicability of the proposed model and answer the following research questions: 

\begin{itemize}[leftmargin=*]
    \item \textbf{RQ1}: Can \ours work well in the node classification task?
    \item \textbf{RQ2}: Can \ours perform well in the graph classification task?
    \item \textbf{RQ3}: Can \ours yield satisfactory results in the transfer learning task on molecular property prediction?
    \item \textbf{RQ4}: What are the performances of \ours \textit{w.r.t.} different model components and decoder backbones?
    \item \textbf{RQ5}: How does \ours perform with various hyper-parameters?
    \item \textbf{RQ6}: Can \ours obtain better representation visualization?
\end{itemize}

\subsection{Experimental Setup}

\noindent
\textbf{Datasets.}
For the node classification task, we employ five public benchmark datasets to evaluate the proposed model, including Cora \cite{cora_citeseer_datasets}, CiteSeer \cite{cora_citeseer_datasets}, PubMed \cite{pubmed}, Ogbn-arxiv \cite{ogb}, and PPI \cite{graphsage}. For the graph classification task, we report the performance on seven benchmark datasets \cite{dgk_and_graph_classification_datasets}: IMDB-M, IMDB-B, PROTEINS, COLLAB, MUTAG, REDDIT-B, and NCI1. For the transfer learning task, we conduct experiments on eight benchmark datasets \cite{molecular_datasets} including BBBP, Tox21, ToxCast, SIDER, ClinTox, MUV, HIV, and BACE.

\noindent
\textbf{Baselines.}
For the node classification task, we compare \ours against supervised models GCN \cite{gcn}, GAT \cite{gat}, contrastive self-supervised models DGI \cite{dgi}, MVGRL \cite{mvgrl}, GRACE \cite{grace}, BGRL \cite{bgrl}, InfoGCL \cite{infogcl}, CCA-SSG \cite{ccassg}, and generative self-supervised models GAE~\cite{gae}, GPT-GNN~\cite{gptgnn}, GATE~\cite{gate}, GraphMAE \cite{graphmae}, and GraphMAE2 \cite{graphmae2}. For the graph classification task, we compare the proposed model with supervised baselines, GIN \cite{gin}, DiffPool \cite{diffpool}, classical graph kernel methods Weisfeiler-Lehman sub-tree kernel (WL) \cite{wl}, deep graph kernel (DGK) \cite{dgk_and_graph_classification_datasets}, as well as contrastive methods GCC \cite{gcc}, graph2vec \cite{graph2vec}, Infograph \cite{infograph}, GraphCL \cite{graphCL}, JOAO \cite{joao},  MVGRL \cite{mvgrl},  InfoGCL \cite{infogcl}, and generative methods GraphMAE \cite{graphmae} and GraphMAE2 \cite{graphmae2}. For the transfer learning task, we evaluate \ours against methods including No-pretrain, ContextPred \cite{hu2020strategies}, AttrMasking \cite{hu2020strategies}, Infomax \cite{infomax}, contrastive methods GraphCL \cite{graphCL}, JOAO \cite{joao}, GraphLoG \cite{graphlog}, as well as generative methods GraphMAE \cite{graphmae} and GraphMAE2 \cite{graphmae2}.

\noindent
\textbf{Implementation Details.}
Baseline results are taken from GraphMAE \cite{graphmae}. The unreported results are due to inaccessible code or memory limitations. For each task, we adhere to the exact same experimental procedure, e.g., data split and evaluation protocol, as the standard settings \cite{dgi, ccassg, infograph, hu2020strategies, graphmae}. 
For the node classification task, the inductive setting is taken from GraphSAGE \cite{graphsage} where the testing is employed on unseen nodes. For the graph classification task, to associate each graph with input features, we utilize node labels as the features for datasets MUTAG, PROTEINS, and NCI1, while leveraging node degrees as the features for datasets IMDB-M, IMDB-B, REDDIT-B, and COLLAB. For the transfer learning task, we follow the experimental setting in \cite{hu2020strategies}. In order to simulate the real-world use cases, the downstream datasets are split using scaffold split. The atom number and chirality tag are employed as node features, which provide information about the properties of individual atoms in a molecule. Additionally, the interactions between atoms in the molecule are represented by using bond type and direction as edge features. The experiments are implemented using PyTorch and DGL library \cite{dgl}. We optimize the model using Adam \cite{adam}.

\begin{table*}[htp]
\caption{Transfer learning performance comparison. We report ROC-AUC scores (\%) for all datasets.
}
\label{tab:transfer_learning}
\renewcommand\arraystretch{1.1}
\resizebox{\linewidth}{!}{
\begin{NiceTabular}{c|cccccccc|c}
\toprule[1.1pt]
Methods & BBBP       & Tox21      & ToxCast    & SIDER      & ClinTox    & MUV        & HIV        & BACE & Avg.      \\
\midrule
    No-pretrain         & 65.5±1.8   & 74.3±0.5   & 63.3±1.5   & 57.2±0.7   & 58.2±2.8   & 71.7±2.3   & 75.4±1.5   & 70.0±2.5 & 67.0  \\
    \cdashlineCustom{1-10}
    ContextPred  \cite{hu2020strategies}       & 64.3±2.8   & 75.7±0.7   & 63.9±0.6   & 60.9±0.6   & 65.9±3.8   & 75.8±1.7   & 77.3±1.0   & 79.6±1.2 & 70.4 \\
    AttrMasking  \cite{hu2020strategies}       & 64.3±2.8   & \bf 76.7±0.4   & 64.2±0.5   &  61.0±0.7   & 71.8±4.1   & 74.7±1.4   & 77.2±1.1   & 79.3±1.6 & 71.1 \\
    Infomax \cite{infomax}            & 68.8±0.8  & 75.3±0.5  & 62.7±0.4  & 58.4±0.8  & 69.9±3.0  & 75.3±2.5  & 76.0±0.7  & 75.9±1.6 & 70.3 \\
    GraphCL  \cite{graphCL}           & 69.7±0.7 & 73.9±0.7 & 62.4±0.6 & 60.5±0.9 & 76.0±2.7 & 69.8±2.7 & \underline{78.5±1.2} & 75.4±1.4 & 70.8 \\
    JOAO  \cite{joao}              & 70.2±1.0 & 75.0±0.3 & 62.9±0.5 & 60.0±0.8 & 81.3±2.5 & 71.7±1.4 & 76.7±1.2 & 77.3±0.5 & 71.9 \\
    GraphLoG  \cite{graphlog}          & \underline{72.5±0.8} &  75.7±0.5  &  63.5±0.7     & \underline{61.2±1.1}  & 76.7±3.3   & 76.0±1.1    & 77.8±0.8  & \underline{83.5±1.2} & 73.4 \\ 
    GraphMAE \cite{graphmae}          & 72.0±0.6 & 75.5±0.6 & 64.1±0.3 & 60.3±1.1 & \underline{82.3±1.2} & 76.3±2.4 & 77.2±1.0 & 83.1±0.9 & \underline{73.8} \\
    GraphMAE2 \cite{graphmae2}          & 71.5±1.5 & 75.8±0.9 & \underline{65.3±0.8} & 59.7±0.6 & 78.9±2.8 & \underline{78.6±1.2} & 76.2±2.3 & 81.2±1.3 & 73.4 \\
    \midrule
    \ours & \bf 72.8$\pm$0.4 & \underline{76.3$\pm$0.3} & \bf 65.6$\pm$0.3 & \bf 64.6$\pm$0.4 & \bf 82.4$\pm$1.3 & \bf 80.0$\pm$0.8 & \bf 79.8$\pm$0.6 & \bf 85.3$\pm$1.0 & \bf 75.9 \\
\bottomrule[1.1pt]
\end{NiceTabular}
}
\end{table*}

\subsection{Node Classification (RQ1)}

We conduct node classification in two scenarios to fully evaluate the model, i.e., transductive and inductive. 
In particular, we conduct experiments on the transductive setting for datasets Cora, CiteSeer, PubMed, and Ogbn-arxiv, while performing experiments on the inductive setting for dataset PPI. In addition, we adhere to the standard experimental settings \cite{mvgrl, graphmae, bgrl} by first training \ours without supervision. Then we freeze the parameters of \ours encoder and obtain the learned node representations. Later, we train a linear classifier on the obtained representations for evaluation. We utilize GAT as the standard encoder/decoder model structure and report the performances of mean and standard deviation for five separate runs with different random seeds. The results are shown in Table \ref{tab:node_classification}. The best and second-best results are highlighted in bold and underlined, respectively. According to the table, we find that our model \ours outperforms all the baselines across all datasets. In particular, traditional generative self-supervised models such as GAE and GPT-GNN perform poorly by simply focusing on the reconstruction of input data, while contrastive self-supervised models such as DGI and InfoGCL perform slightly better by maximizing the agreement between different distributions or augmentations. Among the baselines, the masked autoencoder-based methods GraphMAE and GraphMAE2 have the best results in most cases, demonstrating the success of utilizing the masking strategy in self-supervised learning. However, GraphMAE and GraphMAE2 show several limitations that can significantly affect their learning capabilities. By addressing these issues, \ours achieves the best performance compared to all methods. This validates the significance of addressing the identified limitations and proves the effectiveness of our model.

\subsection{Graph Classification (RQ2)}

We follow the previous experimental settings \cite{graphmae} to use GIN as backbone encoder/decoder models and first train the model without providing any supervision. Then we utilize the encoder and a readout function to obtain the learned graph representations, which are subsequently fed into a downstream LIBSVM \cite{libsvm} classifier for prediction and evaluation. The results are reported in Table \ref{tab:graph_classification}, based on 10-fold cross-validation, with the mean and standard deviation reported for five different runs. As shown in the table, we observe that our model \ours achieves the best performance across various datasets except for a few cases. To illustrate, the dataset MUTAG has the smallest size with only hundreds of graphs, providing the least amount of information to learn, which leads to unstable results with a large variance. In REDDIT-B, the proposed model has a small difference compared to the best baseline GCC, but still remains the second best among all models, and holds a big improvement compared to GraphMAE and GraphMAE2. This again indicates the efficacy of addressing the limitations and the strong learning capability of our model. Furthermore, we can find that \ours can consistently outperform GraphMAE and GraphMAE2 for both mean accuracy and standard deviation. This demonstrates the effectiveness of the proposed model and the property of obtaining stable reconstructions. In addition, considering that the node features of the graphs in these seven datasets are one-hot vectors (i.e., node labels or degrees), the exceptional performance of \ours manifests that the proposed model can learn meaningful representations even when the provided information is insufficient.

\begin{figure}[t]
	\centering
	\includegraphics[width=\columnwidth]{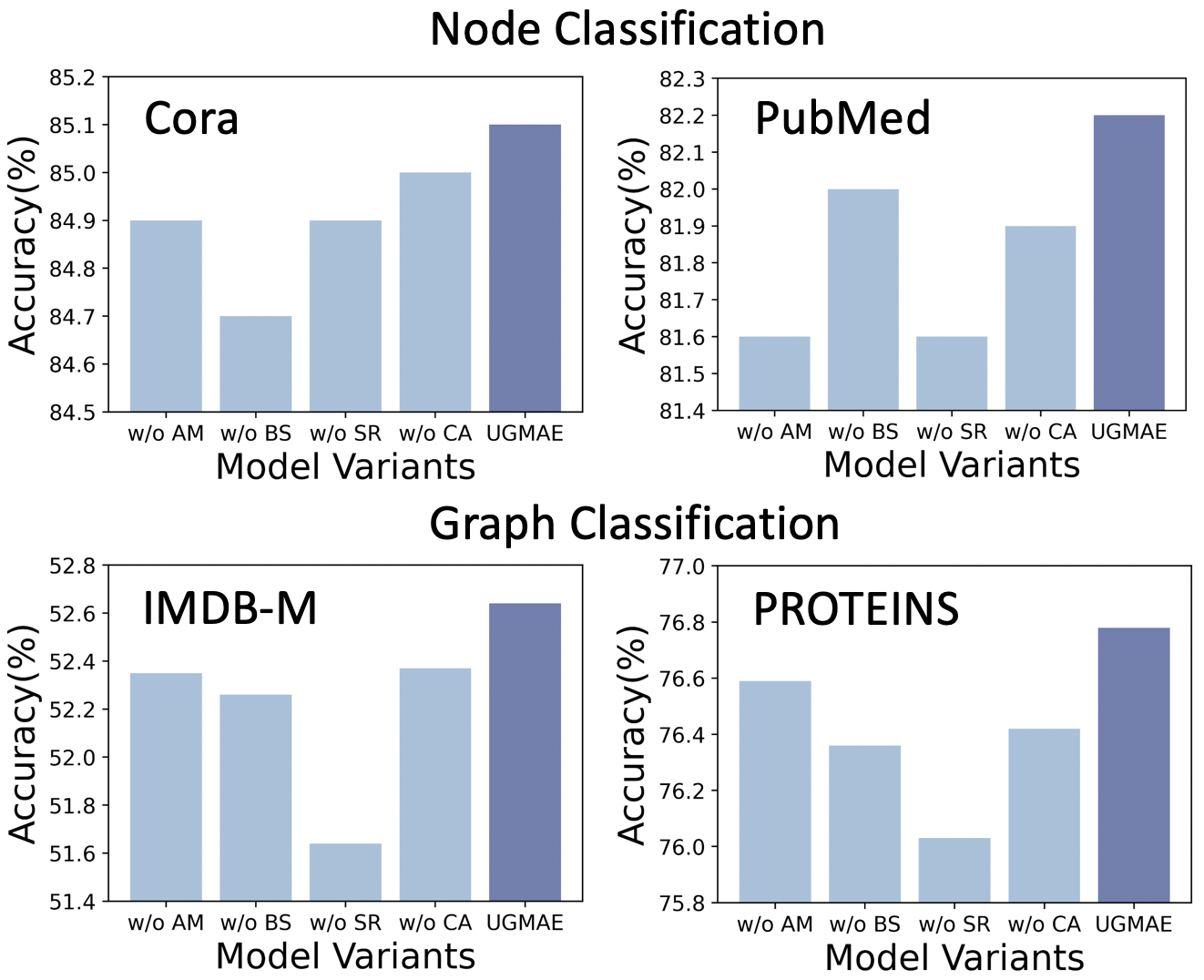}
	\caption{
	Ablation studies of model components.
	}
	\label{fig:ablation}
\end{figure}

\subsection{Transfer Learning (RQ3)}

With the aim of assessing the model transferability, we evaluate the model performance via a transfer learning task on molecular property prediction. In particular, we follow the previous experimental settings \cite{hu2020strategies,graphCL,joao} to first train the model in 2 million unlabeled molecules sampled from ZINC15 \cite{zinc15} and then finetune the model on eight downstream benchmark datasets \cite{molecular_datasets}. We employ a 5-layer GIN as the encoder and a single-layer GIN as the decoder. To fully evaluate the model, we use ROC-AUC scores as the metric and report the mean and standard deviation of 10 individual runs with different random seeds. The results are shown in Table \ref{tab:transfer_learning}. Based on the table, we find that \ours achieves the best results and significantly outperforms baselines in all datasets except in Tox21, in which \ours has the second-best performance. Specifically, although AttrMasking shows the best result in Tox21, it performs poorly on other datasets, with an average ROC-AUC score of 71.1. However, \ours has an average ROC-AUC score of 75.9, which is \textbf{+6.75\%} better than AttrMasking and \textbf{+2.85\%} better than the best baseline GraphMAE, which has an average ROC-AUC score of 73.8. This demonstrates the strong transferability of our model. In addition, we notice that even though GraphMAE has a decent average ROC-AUC score, it can only achieve the best results among baselines on two datasets (i.e., ClinTox and MUV), while our model \ours exhibits a stable major improvement on seven datasets. This again verifies the effectiveness of our model and reflects its incredible learning capacity even in different datasets.

\begin{table}[t]
\caption{
Ablation studies of decoder backbones.
}
\label{tab:ablation}
\renewcommand\arraystretch{1.1}
\renewcommand\tabcolsep{3.5pt}
\begin{center}
\resizebox{\columnwidth}{!}{
\begin{NiceTabular}{ccccccc}
\toprule[1.1pt]
\multicolumn{2}{c}{\multirow{2.3}{*}{Dataset}} & \multicolumn{2}{c}{Node Classification} & \phantom{} & \multicolumn{2}{c}{Graph Classification} \\
\cmidrule{3-4} \cmidrule{6-7} & & Cora & PubMed && IMDB-M & PROTEINS \\

\midrule
\multirow{4}{*}{Decoder}
& MLP & 73.2$\pm$1.3 & 80.3$\pm$0.4 && 51.40$\pm$0.28 & 75.67$\pm$0.37 \\
& GCN & 75.5$\pm$1.9 & 80.4$\pm$0.8 && 51.71$\pm$0.40 & 75.45$\pm$0.42 \\
& GIN & 74.5$\pm$1.6 & 81.4$\pm$0.4 && 52.64$\pm$0.30 & 76.78$\pm$0.22 \\
& GAT & 85.1$\pm$0.4 & 82.2$\pm$0.1 && 50.99$\pm$0.40 & 75.97$\pm$0.47 \\

\bottomrule[1.1pt]
\end{NiceTabular}}
\end{center}
\end{table}

\subsection{Ablation Studies (RQ4)}

\noindent
\textbf{Effect of model components.}
Since \ours contains various model components (i.e., adaptive feature mask generator (AM), ranking-based structure reconstruction (SR), bootstrapping-based similarity module (BS), and consistency assurance module (CA)), we analyze the contributions of different components by removing each of them independently. According to Figure \ref{fig:ablation}, the decreasing performances of removing each component demonstrate the effectiveness of each component in enhancing the model. Specifically, in the node classification, the contributions of each component vary. For the Cora dataset, removing BS significantly affects the performance, showing that BS has a large contribution to \ours. For the PubMed dataset, both AM and SR are the two most important designs in improving the model performance. On the other hand, in graph classification, since the input features of each graph are discrete one-hot encodings constructed using node degrees or node labels, SR plays a crucial role in strengthening the model by enabling it to focus on complex structural information, while other components also contribute. Finally, \ours achieves the best results in all cases, indicating the strong capability of different components in our model.

\noindent
\textbf{Effect of decoder backbones.}
We compare with different decoder backbones such as MLP, GCN, GIN, and GAT to validate the effectiveness of \ours in Table \ref{tab:ablation}. From the results, we observe that the usage of GNNs as the decoder achieves better performances than using MLP in most cases. This might be because of the natural property of GNNs in capturing the complex relational information in graph data via message passing, while MLP simply learns a mapping between the learned representations and the reconstructions. In addition, we notice that using different GNN decoders affects the performance, but the difference is not generally significant across various datasets, except for the Cora dataset. For the node classification, GAT performs the best, while for the graph classification, GIN works the best. Moreover, replacing GAT with any other GNNs in the Cora dataset leads to a substantial performance drop, which is understandable considering Cora has the smallest dataset size. This observation also indicates that the attention mechanism matters in reconstructing continuous node features. Therefore, for the sake of generality and uniformity, we select GAT for all node classification datasets and GIN for all graph classification datasets.

\begin{figure}
	\centering
	\includegraphics[width=\columnwidth]{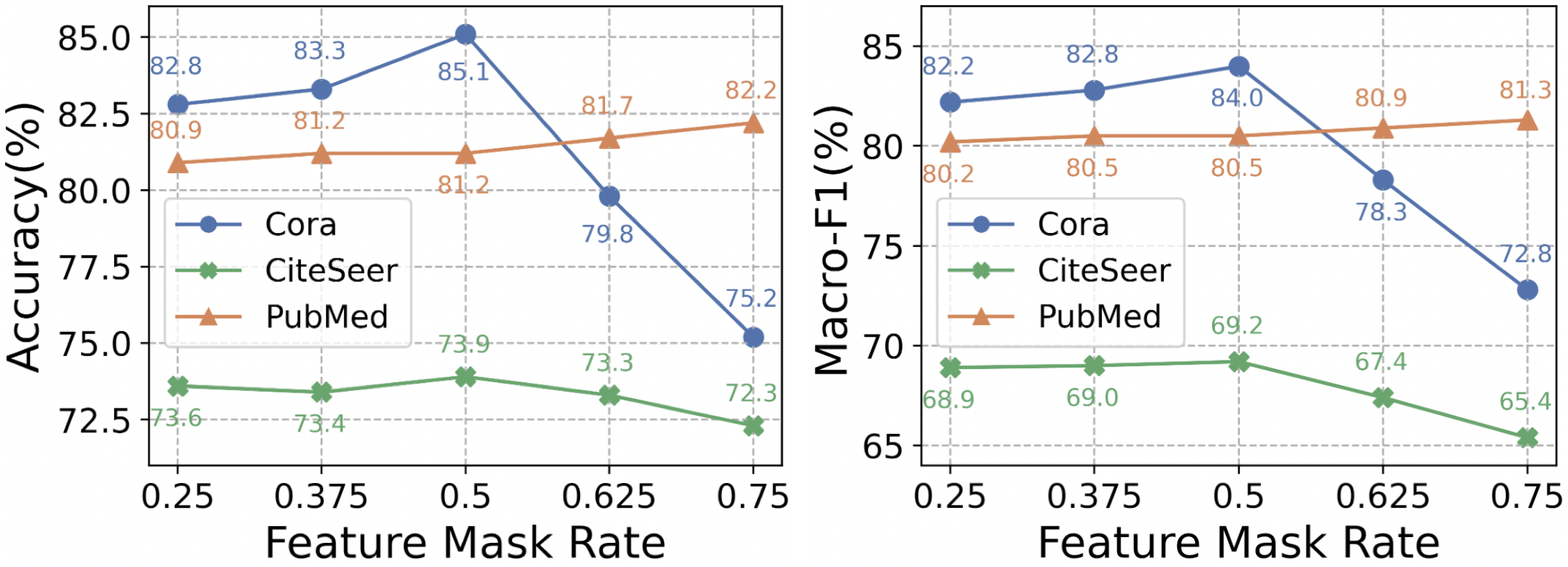}
	\caption{
	The performance of \ours \textit{w.r.t.} different values of feature mask rate $p_f$.
	}
	\label{fig:feature_mask_rate}
\end{figure}

\begin{figure}
	\centering
	\includegraphics[width=\columnwidth]{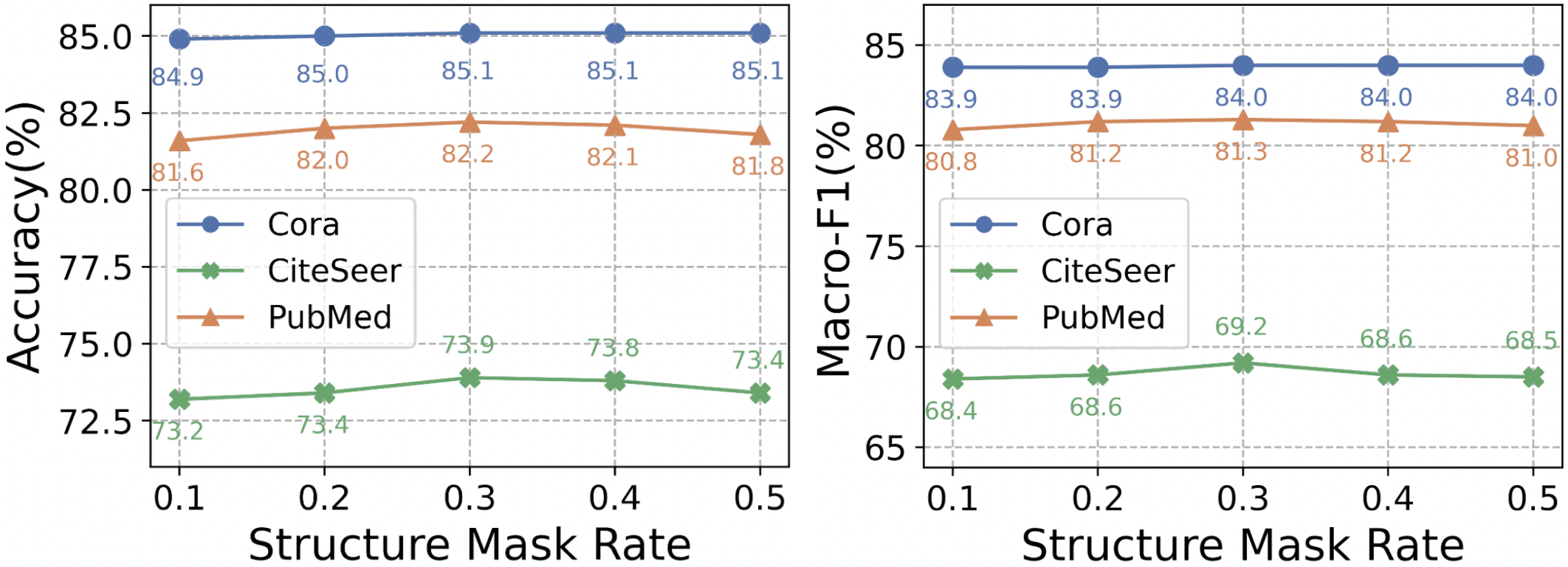}
	\caption{
	The performance of \ours \textit{w.r.t.} different values of structure mask rate $p_s$.
	}
	\label{fig:structure_mask_rate}
\end{figure}

\subsection{Parameter Sensitivity (RQ5)}

\noindent
\textbf{Impact of feature mask rate.}
The feature mask rate $p_f$ controls the percentages of masked nodes. In Figure \ref{fig:feature_mask_rate}, we show the performance of \ours \textit{w.r.t.} different values of feature mask rate $p_f$. In particular, we search the values of $p_f$ from \{0.25, 0.375, 0.5, 0.625, 0.75\}. By analyzing the figure, we find that the optimal feature mask rate can be different across datasets. For example, on Cora and CiteSeer, the model achieves the best performance when $p_f=0.5$, while on PubMed, the model has the highest performance when $p_f=0.9$. Generally, increasing the rate enhances performance. We ascribe this improvement to the comprehensive modeling of the data itself, whereas the use of a low feature mask rate simplifies the learning task and could prevent the model from fully capturing the knowledge. However, further increasing the value could degrade the performance. This is because applying an extreme mask rate provides the model with insufficient information, making it more difficult for the model to encode and concentrate on meaningful knowledge. In addition, the variation in the optimal mask rate for the PubMed dataset compared to other datasets can be attributed to the information redundancy present in the graph structure of the PubMed dataset. This is because the large node degrees or high homogeneity in the PubMed dataset result in high information redundancy, which allows the model to recover node features from a small number of neighbor nodes.

\begin{figure}
	\centering
	\includegraphics[width=\columnwidth]{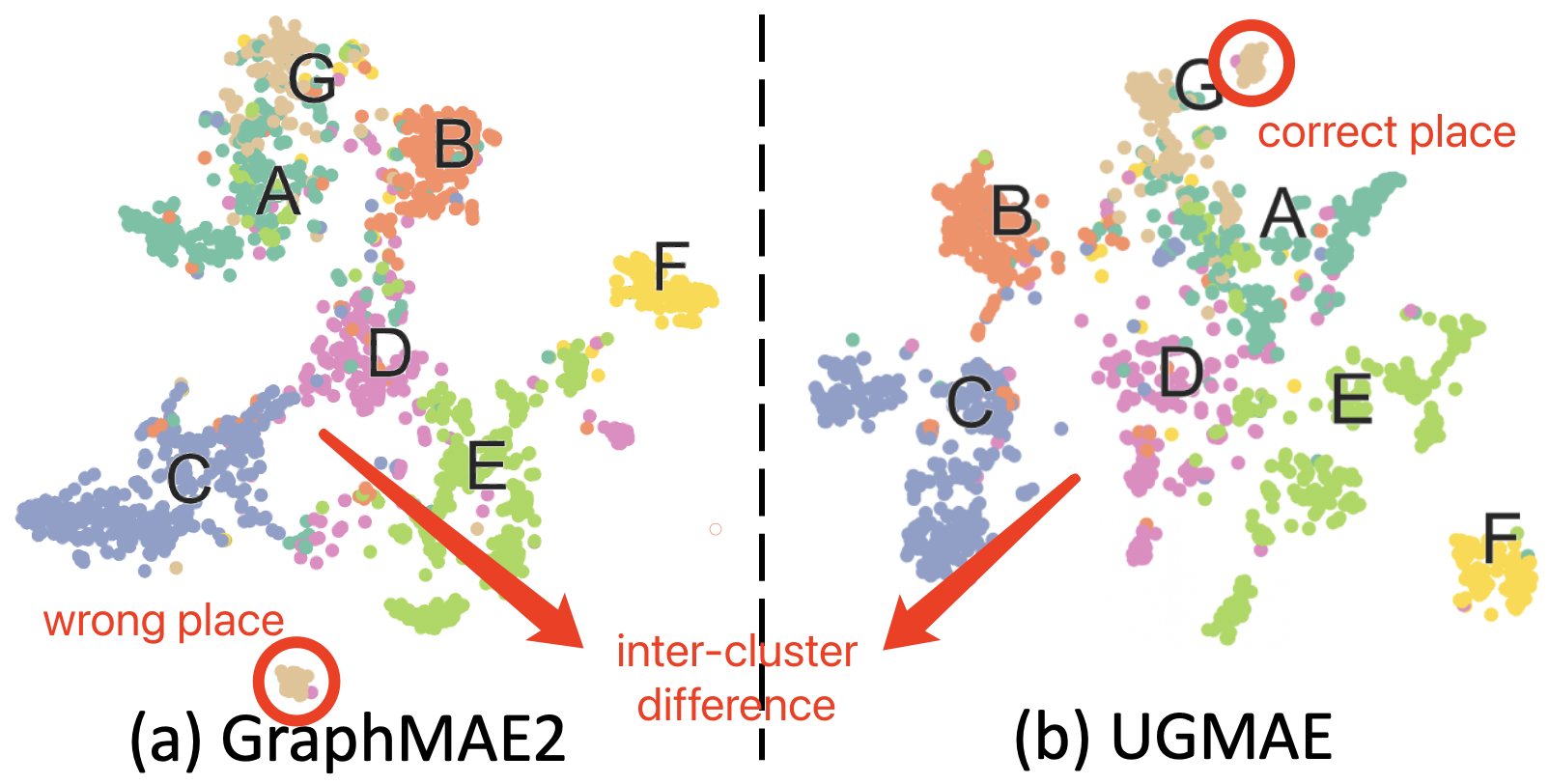}
	\caption{
	Node representation visualization on the Cora dataset. Different colors indicate different node categories.
	}
	\label{fig:visualization}
\end{figure}

\noindent
\textbf{Impact of structure mask rate.}
The structure mask rate $p_s$ controls the percentages of masked edges. Figure \ref{fig:structure_mask_rate} shows the performance of \ours \textit{w.r.t.} different values of structure mask rate $p_s$. Specifically, we search $p_s$ from \{0.1, 0.2, 0.3, 0.4, 0.5\}. From the figure, we observe that the performance reaches the optimal when $p_s=0.3$. For the Cora dataset, the performance stays the same when further increasing the value. However, for other datasets, as the mask rate increases, the performance of the model begins to decrease, although the drop is not significant. In general, the choices of different $p_s$ have little impact on the final model performance, demonstrating the robustness of the proposed model. We thus conclude that \ours is insensitive to the decisions of $p_s$. We ascribe this advantage to the strong learning capacity of model components, which enables \ours to function effectively across various structure mask rates.

\subsection{Visualization (RQ6)}

To gain a more intuitive understanding, we visualize and compare the learned representations of GraphMAE2 and \ours using t-SNE \cite{tsne}. Specifically, we conduct analyses on the Cora dataset and denote the nodes of various categories with different colors. The visualizations are shown in Figure \ref{fig:visualization}. We also mark the seven categories with class names ranging from A to G for a better demonstration. As shown in the figure, we find that GraphMAE2 can successfully recognize the difference between various categories, but the distinction between each category is ambiguous given the clusters are connected to each other. However, \ours can learn discriminative representations that form in separable clusters with a clear inter-cluster difference. For example, for GraphMAE2, G and B are close to each other, and C and D are connected, while \ours can separate G and B well, and divide C and D into two distinct clusters. In addition, \ours can generally learn better clusters than GraphMAE2. To illustrate, G is separated into two chunks for GraphMAE2, one at the very top and the other at the very bottom, whereas \ours can effectively group them. Besides, G is sparse in GraphMAE2 and mixes with E, while \ours can put the nodes of G into a cluster without intertwining with E. This again demonstrates the effectiveness of \ours and the capability of learning discriminative node representations.

\section{Conclusion}

In this paper, we identify and investigate four significant limitations of existing graph masked autoencoder methods.
To address these limitations, We design \ours, a unified framework for graph masked autoencoders from the perspectives of adaptivity, integrity, complementarity, and consistency. Specifically, \ours contains multiple novel components including adaptive feature mask generator, ranking-based structure reconstruction, bootstrapping-based similarity, and consistency assurance. The extensive experiments and in-depth analyses demonstrate the effectiveness and superiority of \ours compared to state-of-the-art methods.

\balance
\bibliography{reference}
\bibliographystyle{ieeetr}

\end{document}